\documentclass[final,5p,times,twocolumn]{elsarticle}

\usepackage{amsmath}
\usepackage{amssymb}
\usepackage{booktabs}
\usepackage{graphicx}
\usepackage{makecell}
\usepackage{multirow}
\usepackage{xspace}
\usepackage{algorithm}
\usepackage{algpseudocode}
\usepackage[table]{xcolor}

\newcommand{\method}{EvLIR\xspace}
\newcommand{\term}{TERM\xspace}
\newcommand{\best}[1]{\textbf{#1}}
\newcommand{\second}[1]{\underline{#1}}

\usepackage[hidelinks]{hyperref}

\makeatletter
\AtBeginDocument{%
  \let\ps@pprintTitle\ps@empty
}
\makeatother

\hypersetup{
    pdfauthor={Haoxian Zhou, Chuanzhi Xu, Langyi Chen, Pengfei Ye, Haodong Chen, Qiang Qu, Ali Anaissi, Ying Zhou, Weidong Cai},
    pdftitle={EvLIR: Learning Illumination Residuals from Ordered Events for Low-Light Image Enhancement}
}

\begin{document}

\begin{frontmatter}

\title{EvLIR: Learning Illumination Residuals from Ordered Events for Low-Light Image Enhancement}

\author[usyd]{Haoxian Zhou\corref{cor1}}
\ead{haoxian.zhou@sydney.edu.au}
\author[usyd]{Chuanzhi Xu\corref{cor1}}
\ead{chuanzhi.xu@sydney.edu.au}
\author[usyd]{Langyi Chen}
\author[mit]{Pengfei Ye}
\author[usyd]{Haodong Chen}

\makeatletter
\g@addto@macro\elsauthors{\par\def\authorsep{}}
\makeatother

\author[usyd]{Qiang Qu}
\author[usyd]{Ali Anaissi}
\author[usyd]{Ying Zhou}
\author[usyd]{Weidong Cai}

\cortext[cor1]{Corresponding author.}

\affiliation[usyd]{
  organization={The University of Sydney},
  city={Sydney},
  state={NSW},
  country={Australia}}

\affiliation[mit]{
  organization={Massachusetts Institute of Technology},
  city={Cambridge},
  state={MA},
  country={USA}}

\begin{abstract}
Low-light image enhancement is severely ill-posed when the input frame contains missing structure, severe noise, and weak local contrast. Event cameras provide asynchronous brightness-change
observations with high temporal resolution, but existing
methods often process event voxels as static channel stacks,
without explicitly modeling their within-window temporal
evolution. We propose EvLIR, a temporal-residual enhancement framework that learns illumination residuals from ordered events for low-light image enhancement.
Given a low-light frame and its aligned event voxel, EvLIR preserves the ordered temporal bins of the event stream and introduces a Temporal Event Residual Module (TERM) to encode short-window event dynamics with a lightweight ConvGRU.
The resulting temporal state is converted into a bounded illumination correction, which provides spatially adaptive photometric guidance for Retinex-style illumination estimation and subsequent reliability-aware image-event restoration.
On SDE and SDSD indoor/outdoor benchmarks, EvLIR achieves the best result on eleven of twelve dataset-metric pairs, with average scores of 25.63 dB PSNR and 0.8271 SSIM across the four benchmarks.
\end{abstract}


\end{frontmatter}

\begin{figure*}[t]
  \centering
  \includegraphics[width=\textwidth]{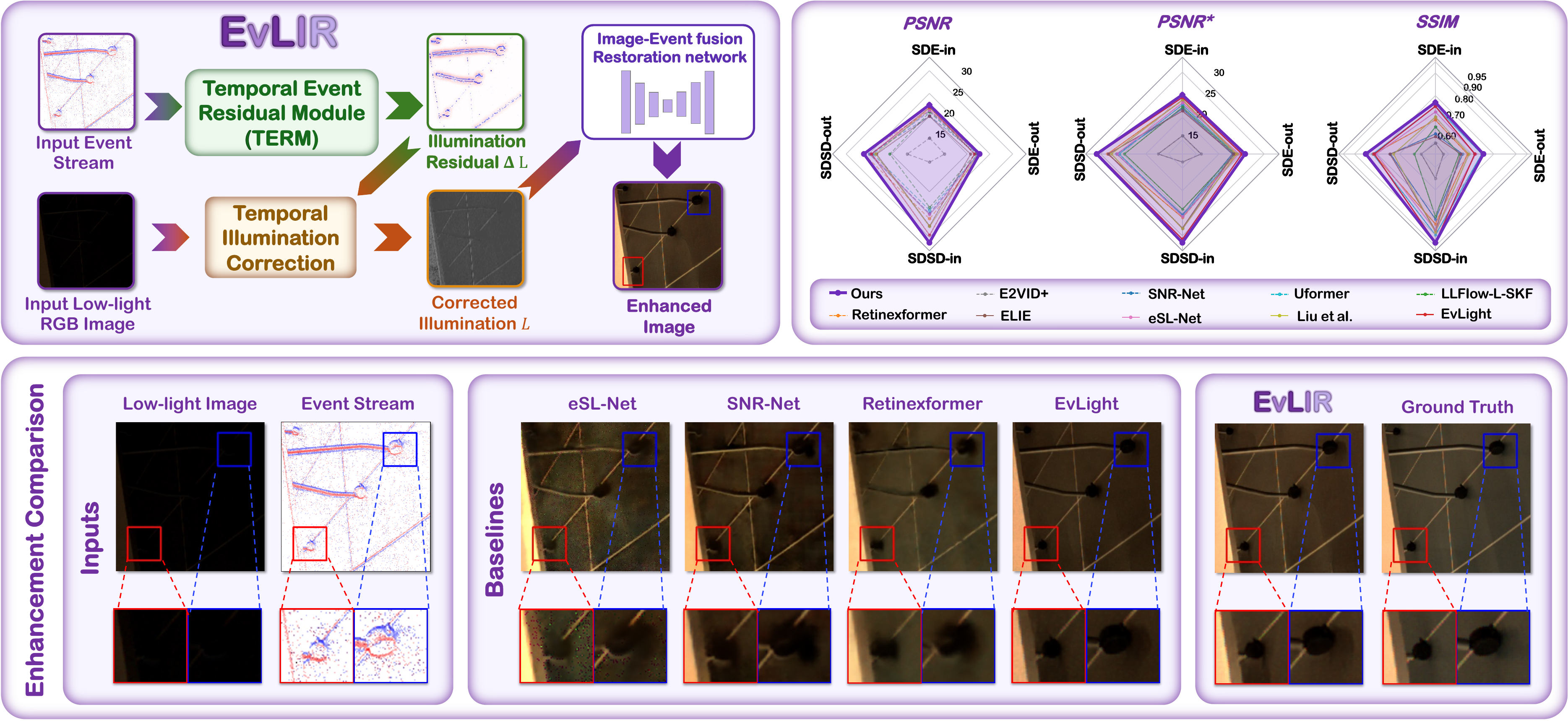}
  \caption{\textbf{Temporal-residual enhancement in \method.}
  A low-light frame provides an unreliable snapshot, while ordered event bins expose short-window brightness-change trends.
  The Temporal Event Residual Module (\term) summarizes these temporal cues and predicts a bounded illumination correction for low-light restoration.}
  \label{fig:teaser}
\end{figure*}

\section{Introduction}
\label{sec:intro}

Low-light image enhancement (LLIE) aims to recover visible, clean, and contrast-rich images from severely under-exposed observations. 
In extreme low light, however, a single frame may not contain enough reliable structure: dark regions are dominated by shot noise, local contrast collapses, and aggressive brightening easily amplifies artifacts~\cite{cnn3}. 
Recent LLIE methods leverage event cameras to provide complementary cues for low-light image enhancement.

Event cameras are bio-inspired sensors that respond asynchronously to brightness changes. They output sparse events with timestamp, spatial location, and polarity, instead of dense frames at a fixed rate~\cite{intro2, chen2026fracevent}.
Compared with conventional cameras, event cameras provide microsecond-level temporal resolution, low latency, high dynamic range, and reduced motion blur. 
These advantages have enabled their use in various vision tasks, such as overexposure reconstruction~\cite{intro4-3}, deblurring~\cite{intro4-1}, occlusion removal~\cite{intro4-2}, super-resolution~\cite{intro5}, and 3D reconstruction~\cite{intro6, 11210052}. 
In particular, their high dynamic range and sensitivity to brightness changes make them well suited for challenging illumination scenarios. Some methods have explored using events to guide low-light image enhancement~\cite{LLEE18,LLEE17,LLEE7,LLEE6}.

For LLIE, the temporal sequence of events can provide useful information. Events occurring within a short time window are more likely to reflect consistent brightness changes or structural cues, while isolated events may be less reliable and can be caused by sensor noise or transient motion~\cite{intro3, intro7}. 
Based on this observation, we hypothesize that a low-light enhancer should consider not only the spatial presence of events, but also their short-term temporal consistency when estimating illumination.

In this paper, we propose Event-based Illumination Residual Learning (EvLIR), a temporal-residual enhancement framework for event-guided low-light image enhancement. Instead of treating event voxels as static multi-channel inputs, \method models their short-window temporal evolution and uses it to guide illumination estimation, as summarized in Fig.~\ref{fig:teaser}.
Specifically, a 32-channel event voxel is divided into $K$ ordered groups, where each group is encoded by a lightweight convolutional projector. A ConvGRU is then used to aggregate the ordered event features and capture intra-window temporal dependencies. 
Within the Temporal Event Residual Module (\term), the final
hidden state is mapped to a bounded illumination correction,
providing spatially adaptive photometric guidance before
image-event feature fusion.

The bounded residual design is important for event-guided LLIE, since event signals can be sparse, noisy, and highly dependent on scene content.
By injecting event-derived temporal corrections into the illumination path, \method couples short-term brightness-change trends with image-event restoration and provides spatially adaptive photometric guidance.

Our contributions are summarized as follows.
\begin{itemize}
    \item We formulate event-guided low-light enhancement through
ordered temporal illumination correction, connecting
within-window brightness-change dynamics to spatially
adaptive enhancement gains before image-event restoration.
    \item We design a Temporal Event Residual Module (\term) that sequentially aggregates ordered event bins and predicts
a bounded residual to refine image-derived illumination. The module transforms asynchronous event dynamics into spatially adaptive photometric guidance, enabling the enhancer to refine low-light illumination according to short-term brightness-change trends. Controlled studies examine chronological order, event-path
complementarity, and temporal-feature placement.
    \item We present a unified temporal-residual enhancement framework that couples ordered event modeling, illumination correction, and reliability-aware image-event restoration. Experiments on SDE and SDSD datasets demonstrate strong and consistent performance.
\end{itemize}

\section{Related Work}
\label{sec:related}

\subsection{Frame-Based Low-Light Image Enhancement}
Low-light image enhancement aims to improve the brightness, contrast, and detail visibility of images captured under insufficient illumination.
Traditional frame-based methods usually rely on hand-crafted priors, such as histogram equalization~\cite{z1,z2,z3} and Retinex theory~\cite{ret1,ret2,ret3,ret4,ret5}.
However, these methods depend on strong assumptions about image formation and often struggle in complex real-world scenes.

Learning-based methods substantially improve LLIE performance under realistic and complex illumination conditions.
Early CNN-based approaches, including LLNet~\cite{cnn1}, RetinexNet~\cite{cnn2}, and KinD~\cite{cnn3}, study low-light restoration from the perspectives of end-to-end enhancement, Retinex decomposition, and layer-wise image restoration.
Transformer-based methods, such as LLFormer~\cite{trans1} and Retinexformer~\cite{trans2}, further exploit long-range dependency modeling to improve detail reconstruction and global consistency.
More recently, Bai et al. proposed DRWKV~\cite{rwkv}, an RWKV-based LLIE model that improves edge continuity with relatively low computational cost.
Most supervised deep LLIE models rely on paired low-light and normal-light images, which are difficult to collect at scale in real scenes.
To address this problem, unpaired or no-reference methods such as EnlightenGAN~\cite{sc1} have been proposed.
More recent generative models, such as Diff-Retinex~\cite{sc2}, further improve perceptual quality and detail recovery using diffusion priors.
However, these models often introduce higher model complexity and inference cost.

Despite their progress, frame-based methods remain constrained by the imaging physics of RGB sensors.
Under extremely low illumination, a large amount of visual information may already be lost during acquisition.
Short exposure produces severe noise, while long exposure can introduce motion blur.
Once scene structures and textures are lost during image capture, image enhancement alone can hardly recover them reliably.

\subsection{Event-Guided Low-Light Image Enhancement}
Event cameras provide high dynamic range, high temporal resolution, and reduced motion blur, making them useful for low-light perception.
Early event-based methods related to low-light imaging mainly focus on reconstructing images or videos from events alone, such as~\cite{event1,event2}.
These methods show the potential of events as an alternative visual signal, but they do not fully exploit the complementary relationship between a low-light RGB frame and its aligned event stream.

Recent works explore the joint use of low-light RGB images and events for LLIE, where events provide complementary high-contrast structural cues under poor illumination. Early studies mainly use events to compensate for missing structures in low-light frames. For example, Zhang et al.~\cite{LLEE18} formulate low-light event-to-image reconstruction as an unsupervised domain adaptation problem, while Jin et al.~\cite{LLEE17} reconstruct event gradients and fuse them with low-light image features through a dual-branch GAN. Subsequent methods further improve cross-modal fusion and illumination modeling. Jiang et al.~\cite{LLEE7} combine event streams and under-exposed frames with residual fusion, multi-scale Transformer encoding, and multi-level reconstruction losses. Wu et al.~\cite{LLEE5} propose an unsupervised event-assisted enhancement method by matching simulated event maps with real events. More recent methods incorporate Retinex-style illumination estimation and controllable enhancement. Sun et al.~\cite{LLEE4} and Guo et al.~\cite{LLEE3} use event information to guide illumination estimation and Retinex decomposition, while Lu et al.~\cite{LLEE2} introduce brightness prompts for controllable low-light enhancement. Liang et al.~\cite{LLEE6} propose EvLight and construct SDE, a large-scale real-world event-image dataset with spatio-temporal alignment. SDE establishes a representative benchmark for event-guided LLIE and enables systematic evaluation under indoor and outdoor low-light scenes.

Although these methods demonstrate the value of events for LLIE, most of them still treat events mainly as structural edges, illumination priors, or static voxel features.
The temporal order inside the event window is usually compressed before fusion, which makes it harder to distinguish stable brightness changes from noisy event responses.
Recurrent event processing has also been explored in related
restoration tasks. REFID~\cite{Sun_2023_CVPR} uses bidirectional
event recurrence for frame interpolation with ad-hoc deblurring.
In \method, recurrence operates over temporal groups within
a frame-aligned event window, and the resulting state predicts
a bounded correction to image-derived illumination.
This design connects ordered brightness-change observations
to spatial enhancement gains before image-event restoration.
The feature-placement study further evaluates this photometric
route against direct temporal-feature fusion.

\begin{figure*}[t]
    \centering
    \includegraphics[width=\textwidth]{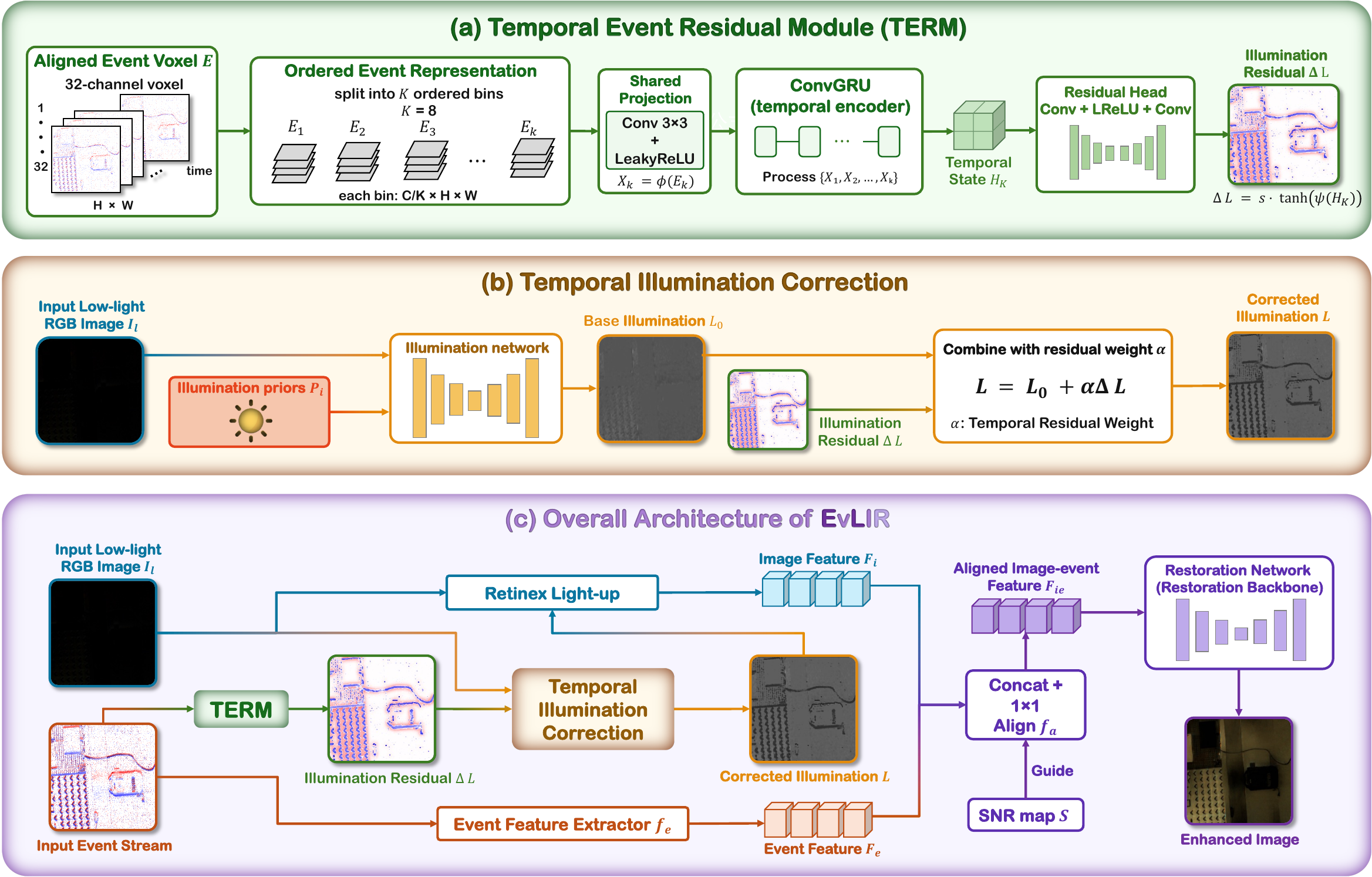}
    \caption{\textbf{Architecture of \method.} Ordered event bins are processed by the Temporal Event Residual Module to predict a bounded illumination correction. The corrected illumination drives Retinex light-up, and reliability-aware image-event restoration produces the final image.}
    \label{fig:overview}
\end{figure*}

\section{Method}
\label{sec:method}

\subsection{Base Illumination Estimation}

Given a low-light RGB image $I_l \in \mathbb{R}^{3 \times H \times W}$
and an aligned event voxel $E \in \mathbb{R}^{C \times H \times W}$,
\method aims to recover a normal-light image $\hat{I}$.
We formulate the enhancement process through Retinex-style illumination
estimation. An illumination network first estimates a base illumination map
$L_0 \in \mathbb{R}^{1 \times H \times W}$ from the low-light observation
and its illumination prior:
\begin{equation}
    L_0, F_l = \mathcal{I}(I_l, P_i),
    \label{eq:base_illumination}
\end{equation}
where $P_i$ denotes the illumination prior and $F_l$ is the illumination
feature. Specifically, $P_i$ is computed directly from the low-light RGB
image using the per-pixel Max-RGB operation:
\begin{equation}
    P_i(x) = \max_{c \in \{R,G,B\}} I_l^c(x),
    \label{eq:illumination_prior}
\end{equation}
where $I_l^c(x)$ denotes the intensity of channel $c$ at spatial location
$x$. This construction follows the illumination-estimation principle in
LIME~\cite{ret1} and uses only the observed low-light image, without accessing
the corresponding normal-light target or any ground-truth illumination
information.

The base illumination map provides the initial photometric estimate for
low-light correction. Here, $L_0$ serves as an illumination-guided spatial
enhancement correction rather than an estimate of the absolute physical
illumination of the scene. Instead of using this estimate directly, \method
refines it with temporal information from ordered events. The overall
architecture is shown in Fig.~\ref{fig:overview}.

\subsection{Ordered Event Representation}

The event voxel contains $C$ channels produced by temporal discretization.
Directly treating these channels as a static feature map can obscure the short-window order of brightness changes.
To preserve this temporal structure, we divide the event voxel into $K$ ordered bins:
\begin{equation}
    E = [E_1, E_2, \ldots, E_K], \qquad
    E_k \in \mathbb{R}^{\frac{C}{K} \times H \times W}.
    \label{eq:event_bins}
\end{equation}
Each $E_k$ represents one ordered time segment in the event window.
This representation allows the network to distinguish consistent brightness-change trends from isolated event responses and provides the input sequence for the temporal residual estimation.

\subsection{Temporal Event Residual Module}

The Temporal Event Residual Module (\term) converts the ordered event bins into a temporal feature for illumination correction.
For each event bin, \term first applies a $3\times3$ convolution followed by LeakyReLU:
\begin{equation}
    X_k = \phi(E_k),
    \label{eq:event_projection}
\end{equation}
where $\phi$ denotes the convolutional projector.
The projected sequence is then encoded by a ConvGRU.
For the $k$-th bin, the recurrent update is represented as:
\begin{align}
    Z_k &= \sigma(W_z * [X_k, H_{k-1}]), \\
    R_k &= \sigma(W_r * [X_k, H_{k-1}]), \\
    \tilde{H}_k &= \tanh(W_h * [X_k, R_k \odot H_{k-1}]), \\
    H_k &= (1 - Z_k) \odot H_{k-1} + Z_k \odot \tilde{H}_k,
    \label{eq:convgru}
\end{align}
where $H_k$ is the hidden state, $Z_k$ and $R_k$ are the update and reset gates, $*$ denotes convolution, and all ConvGRU gates use $3\times3$ convolutions.
The temporal state $H_k$ after the last ordered bin summarizes the short-term event dynamics inside the window.

This sequential design is important because ordered event bins should not be reduced to a static channel stack too early.
A direct concatenation of all bins can mix temporal and spatial evidence, but it does not explicitly model how brightness changes evolve from early to late bins.
Similarly, averaging the per-bin features preserves the existence of multiple bins but removes their temporal transition, making persistent changes and isolated event responses harder to separate.
In \term, the ConvGRU processes the bins in order and updates the hidden state through reset and update gates.
Consistent brightness-change patterns can be reinforced across consecutive bins, while sparse or transient responses can be down-weighted when they are not supported by the temporal context.
The resulting state predicts a spatially adaptive illumination correction, allowing different regions to receive residual adjustments based on local event dynamics.
The pseudo-code of \term is given in Algorithm~\ref{alg:term_pseudocode}.

\definecolor{LightPurple}{RGB}{238,232,248}

\begin{table*}[t]
\centering
\scriptsize
\caption{Quantitative comparison on SDE and SDSD benchmarks. The best and second-best results in each metric are shown in bold and underlined, respectively.}
\label{tab:main_results}
\resizebox{\textwidth}{!}{
\begin{tabular}{llccc ccc ccc ccc}
\toprule
\multirow{2}{*}{Input} & \multirow{2}{*}{Method}
& \multicolumn{3}{c}{SDE-in}
& \multicolumn{3}{c}{SDE-out}
& \multicolumn{3}{c}{SDSD-in}
& \multicolumn{3}{c}{SDSD-out} \\
\cmidrule(lr){3-5}\cmidrule(lr){6-8}\cmidrule(lr){9-11}\cmidrule(lr){12-14}
& & PSNR & PSNR* & SSIM & PSNR & PSNR* & SSIM & PSNR & PSNR* & SSIM & PSNR & PSNR* & SSIM \\
\midrule
Event only & E2VID+~\cite{event2}
& 15.19 & 15.92 & 0.5891 & 15.01 & 16.02 & 0.5765 & 13.48 & 13.67 & 0.6494 & 16.58 & 17.27 & 0.6036 \\
\midrule
Image only & SNR-Net~\cite{snr}
& 20.05 & 21.89 & 0.6302 & 22.18 & 22.93 & 0.6611 & 24.74 & 25.30 & 0.8301 & 24.82 & 26.44 & 0.7401 \\
Image only & Uformer~\cite{uformer}
& 21.09 & 22.75 & 0.7524 & 22.32 & 23.57 & \second{0.7469} & 24.03 & 25.59 & 0.8999 & 24.08 & 25.89 & \second{0.8184} \\
Image only & LLFlow-L-SKF~\cite{skf}
& 20.92 & 22.22 & 0.6610 & 21.68 & 23.41 & 0.6467 & 23.39 & 24.13 & 0.8180 & 20.39 & 24.73 & 0.6338 \\
Image only & Retinexformer~\cite{trans2}
& 21.30 & 23.78 & 0.6920 & \best{22.92} & 23.71 & 0.6834 & 25.90 & 25.97 & 0.8515 & \second{26.08} & 28.48 & 0.8150 \\
\midrule
Image+Event & ELIE~\cite{LLEE7}
& 19.98 & 21.40 & 0.6178 & 20.69 & 23.12 & 0.6533 & 27.46 & 28.30 & 0.8793 & 23.29 & 28.26 & 0.7423 \\
Image+Event & eSL-Net~\cite{esl}
& 21.25 & 23.19 & 0.7277 & 22.42 & 24.39 & 0.7187 & 24.99 & 25.72 & 0.8786 & 24.49 & 26.36 & 0.8031 \\
Image+Event & Liu et al.~\cite{liu}
& 21.79 & 23.88 & 0.7051 & 22.35 & 23.89 & 0.6895 & 27.58 & 28.43 & 0.8879 & 23.51 & 27.63 & 0.7263 \\
Image+Event & EvLight~\cite{LLEE6}
& \second{21.98} & \second{24.41} & \second{0.7527} & 21.49 & \second{25.08} & 0.7159 & \second{29.47} & \second{30.77} & \second{0.9341} & 24.39 & \second{30.67} & 0.8123 \\
\midrule
\rowcolor{LightPurple}
Image+Event & \method (ours)
& \best{22.50} & \best{25.00} & \best{0.7699} & \second{22.72} & \best{25.78} & \best{0.7540} & \best{31.18} & \best{31.53} & \best{0.9346} & \best{26.10} & \best{30.89} & \best{0.8498} \\
\bottomrule
\end{tabular}}
\end{table*}

\begin{algorithm}[t]
  \caption{Temporal Event Residual Module in \method}
  \label{alg:term_pseudocode}
  \begin{algorithmic}[1]
    \Statex \textbf{Input:} event voxel $E$, base illumination $L_0$, bin number $K$
    \Statex \textbf{Output:} corrected illumination $L$
    \State Split $E$ into ordered bins $[E_1,E_2,\ldots,E_K]$
    \State Initialize temporal state $H_0\gets\mathbf{0}$
    \For{$k=1,\ldots,K$}
      \State Project current bin: $X_k\gets\phi(E_k)$
      \State Form recurrent input: $U_k\gets[X_k,H_{k-1}]$
      \State Update gate: $Z_k\gets\sigma(W_z * U_k)$
      \State Reset gate: $R_k\gets\sigma(W_r * U_k)$
      \State Candidate input: $V_k\gets[X_k,R_k\odot H_{k-1}]$
      \State Candidate state: $\tilde{H}_k\gets\tanh(W_h * V_k)$
      \State $H_k\gets(1-Z_k)\odot H_{k-1}+Z_k\odot\tilde{H}_k$
    \EndFor
    \State Predict residual: $\Delta L\gets s\cdot\tanh(\psi(H_k))$
    \State Refine illumination: $L\gets L_0+\alpha\Delta L$
    \State \Return $L$
  \end{algorithmic}
\end{algorithm}

\subsection{Temporal Illumination Correction}

\term converts the temporal state $H_k$ into a bounded illumination residual.
The residual head $\psi$ consists of a $3\times3$ convolution, LeakyReLU, and a final $3\times3$ convolution:
\begin{equation}
    \Delta L = s \cdot \tanh(\psi(H_k)),
    \label{eq:bounded_residual}
\end{equation}
where $H_k$ denotes the state after the last ordered bin has been processed, and $s$ controls the residual scale.
The corrected illumination map is then computed as:
\begin{equation}
    L = L_0 + \alpha \Delta L,
    \label{eq:illumination_update}
\end{equation}
where $\alpha$ controls the contribution of the temporal residual.
Here, $L$ is an illumination-guided spatial correction. The bounded residual transforms asynchronous event dynamics into spatially adaptive photometric guidance, enabling the enhancer to refine low-light illumination according to short-term brightness-change trends.
The corrected illumination is used for Retinex light-up:
\begin{equation}
    I_m = I_l \odot L + I_l.
    \label{eq:retinex_lightup}
\end{equation}
Equivalently, the spatial enhancement gain is $G=1+L$.
The temporal residual adjusts this gain according to
event dynamics, with
$|L(x)-L_0(x)|\leq \alpha s$ at each spatial location.
This bounded update allows temporal evidence to refine
local enhancement strength while retaining the
image-derived base estimate.

\subsection{Reliability-Aware Image-Event Restoration}

After temporal illumination correction, \method extracts image and event features for restoration.
The image branch encodes the light-up image $I_m$, while the event branch encodes the original event voxel:
\begin{equation}
    F_i = f_i(I_m), \qquad F_e = f_e(E).
\end{equation}
The two feature maps are concatenated and aligned by a $1\times1$ convolution:
\begin{equation}
    F_{ie} = f_a([F_i, F_e]).
\end{equation}
An SNR map provides a spatial reliability prior for image-event restoration~\cite{LLEE6}.
High-SNR regions indicate more reliable image evidence, whereas low-SNR regions encourage stronger use of event features.
The restoration network takes the aligned image-event feature $F_{ie}$, the light-up image $I_m$, image feature $F_i$, event feature $F_e$, and SNR map as inputs to predict the final normal-light image:
\begin{equation}
    \hat{I} = \mathcal{R}(F_{ie}, I_m, F_i, F_e, S).
\end{equation}
In this way, \method jointly organizes ordered event dynamics, illumination correction, and spatially adaptive modality selection within a temporal-residual enhancement framework.

\section{Experiments and Results}
\label{sec:experiments}

\begin{figure*}[t]
\centering
\includegraphics[width=\textwidth]{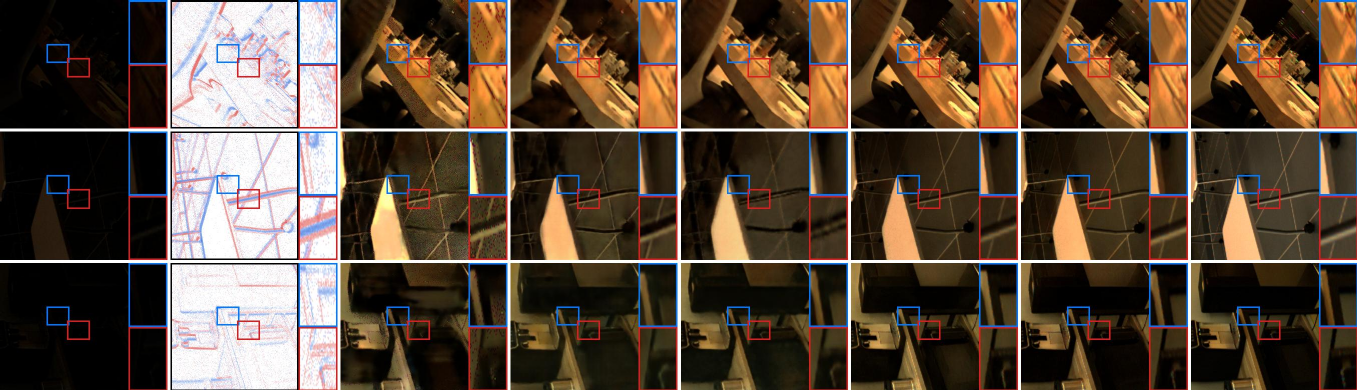}\\[-0.15em]
{\scriptsize
\makebox[0.125\textwidth][c]{Input}\makebox[0.125\textwidth][c]{Event}\makebox[0.125\textwidth][c]{eSL-Net}\makebox[0.125\textwidth][c]{SNR-Net}\makebox[0.125\textwidth][c]{Retinexformer}\makebox[0.125\textwidth][c]{EvLight}\makebox[0.125\textwidth][c]{Ours}\makebox[0.125\textwidth][c]{GT}
}
\caption{Qualitative comparison on SDE-in.}
\label{fig:qual_sde_in}
\end{figure*}

\begin{figure*}[t]
\centering
\includegraphics[width=\textwidth]{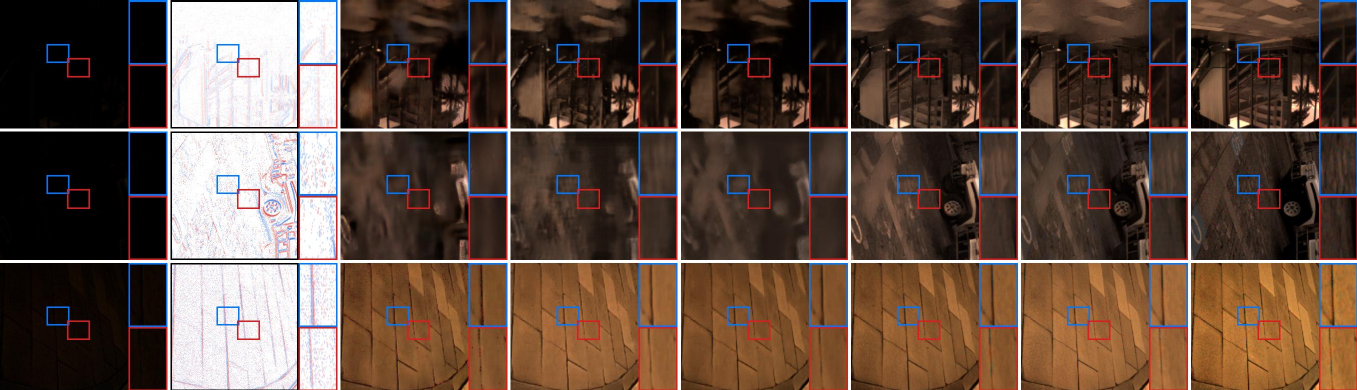}\\[-0.15em]
{\scriptsize
\makebox[0.125\textwidth][c]{Input}\makebox[0.125\textwidth][c]{Event}\makebox[0.125\textwidth][c]{eSL-Net}\makebox[0.125\textwidth][c]{SNR-Net}\makebox[0.125\textwidth][c]{Retinexformer}\makebox[0.125\textwidth][c]{EvLight}\makebox[0.125\textwidth][c]{Ours}\makebox[0.125\textwidth][c]{GT}
}
\caption{Qualitative comparison on SDE-out.}
\label{fig:qual_sde_out}
\end{figure*}

\begin{figure*}[t]
\centering
\includegraphics[width=\textwidth]{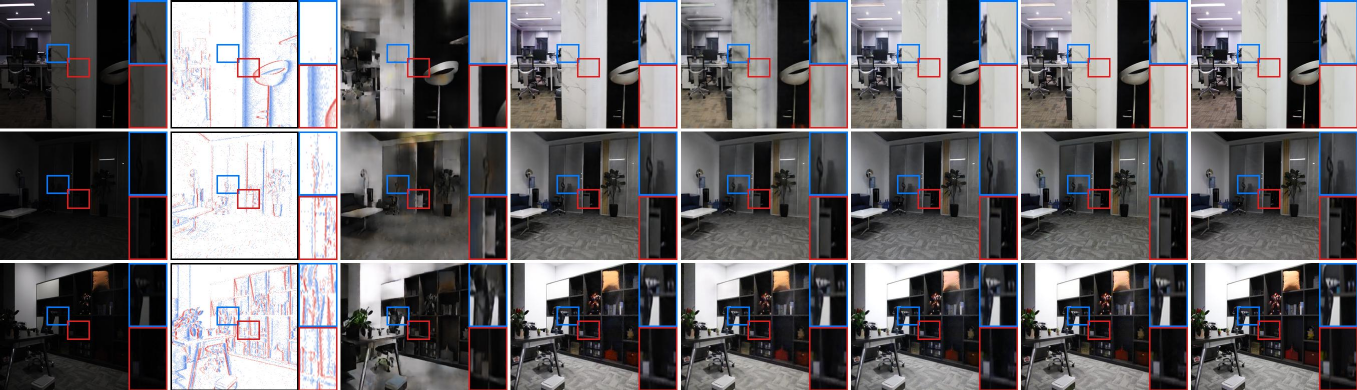}\\[-0.15em]
{\scriptsize
\makebox[0.125\textwidth][c]{Input}\makebox[0.125\textwidth][c]{Event}\makebox[0.125\textwidth][c]{eSL-Net}\makebox[0.125\textwidth][c]{SNR-Net}\makebox[0.125\textwidth][c]{Retinexformer}\makebox[0.125\textwidth][c]{EvLight}\makebox[0.125\textwidth][c]{Ours}\makebox[0.125\textwidth][c]{GT}
}
\caption{Qualitative comparison on SDSD-in.}
\label{fig:qual_sdsd_in}
\end{figure*}

\begin{figure*}[t]
\centering
\includegraphics[width=\textwidth]{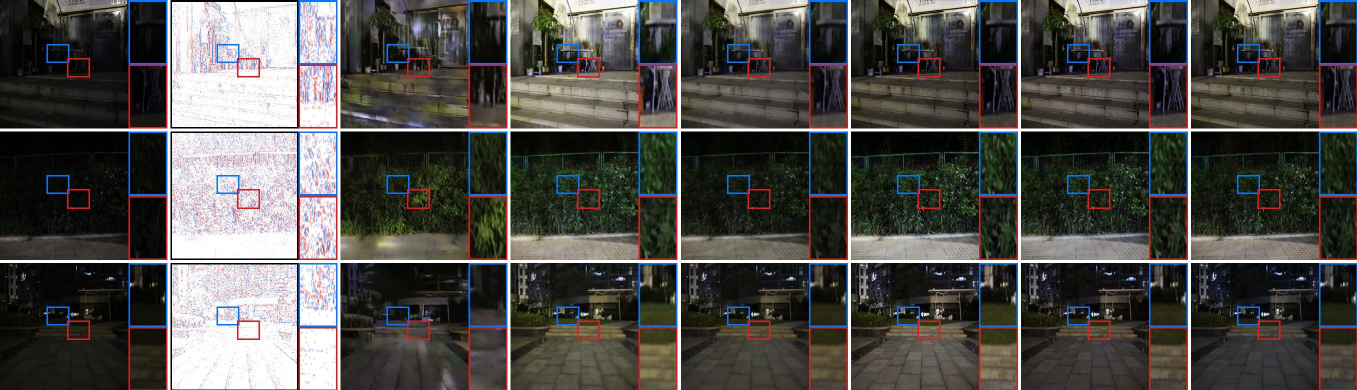}\\[-0.15em]
{\scriptsize
\makebox[0.125\textwidth][c]{Input}\makebox[0.125\textwidth][c]{Event}\makebox[0.125\textwidth][c]{eSL-Net}\makebox[0.125\textwidth][c]{SNR-Net}\makebox[0.125\textwidth][c]{Retinexformer}\makebox[0.125\textwidth][c]{EvLight}\makebox[0.125\textwidth][c]{Ours}\makebox[0.125\textwidth][c]{GT}
}
\caption{Qualitative comparison on SDSD-out.}
\label{fig:qual_sdsd_out}
\end{figure*}

\subsection{Experimental Setup}

\paragraph{Datasets}
We evaluate on four event-guided LLIE benchmarks: SDE-in, SDE-out, SDSD-in, and SDSD-out.
The SDE~\cite{LLEE6} dataset contains 91 paired image-event sequences, including 43 indoor and 48 outdoor sequences, with 76 sequences for training and 15 for testing. All data are captured at a resolution of $346\times260$.
The SDSD~\cite{sdsd} benchmark contains paired low-light and normal-light videos from dynamic indoor and outdoor scenes at an original resolution of $1920\times1080$. Following the dynamic-subset protocol, 125 sequences are used for training and 25 for testing. The videos are resized to $346\times260$, and synthetic events are generated from the low-light sequences using v2e; the normal-light videos are used only as reconstruction targets.

\paragraph{Benchmarks and Metrics}
We compare \method with representatively leading and reproducible baselines, including event-only, image-only, and image-event enhancement methods.
E2VID+ uses only the event stream, while SNR-Net, Uformer, LLFlow-L-SKF, and Retinexformer take only the low-light RGB image as input.
ELIE, eSL-Net, Liu et al., EvLight, and \method use both the low-light image and the aligned event representation.
We report Peak Signal-to-Noise Ratio (PSNR)~\cite{psnr}, PSNR*~\cite{LLEE6}, and the Structural Similarity Index Measure (SSIM)~\cite{ssim} as image restoration metrics.

\paragraph{Implementation Details}
All images are evaluated at $260\times346$ resolution.
For \method, the event voxel has 32 channels and is split into $K=8$ ordered temporal bins by default.
The temporal hidden dimension is 16, with $\alpha=0.1$ and residual scale $s=0.1$.
We train the models with Adam using an initial learning rate of $1.5\times10^{-4}$ and a cosine learning-rate schedule for 80 epochs.
Training and evaluation are conducted on a single NVIDIA RTX 5090 GPU. During training, the GPU memory usage is approximately 60\%. The training time is about 8 hours on SDE-in, 7.5 hours on SDE-out, 8.5 hours on SDSD-in, and 10 hours on SDSD-out.

\subsection{Quantitative Comparison}

Table~\ref{tab:main_results} shows that \method achieves the best result on eleven of the twelve dataset-metric pairs.
On SDE-in, \method reaches 22.50 dB PSNR, 25.00 dB PSNR*, and 0.7699 SSIM, improving the best competing results by 0.52 dB, 0.59 dB, and 0.0172, respectively.
On SDE-out, \method obtains the highest PSNR* and SSIM, with gains of 0.70 dB and 0.0071 over the best competing scores.
On SDSD-in, \method ranks first on all three metrics, including a 1.71 dB PSNR improvement over the best competing result, indicating that the temporal cue remains useful when events are generated synthetically.
On SDSD-out, it achieves 26.10 dB PSNR, 30.89 dB PSNR*, and 0.8498 SSIM, surpassing the previous best results by 0.02 dB, 0.22 dB, and 0.0314.

Averaged across the four benchmarks, \method achieves
25.63 dB PSNR, 28.30 dB PSNR*, and 0.8271 SSIM.
Although Retinexformer obtains higher raw PSNR on SDE-out
(22.92 versus 22.72 dB), \method achieves the best PSNR*
and SSIM on this subset.

\subsection{Qualitative Comparison}

Figures~\ref{fig:qual_sde_in}-\ref{fig:qual_sdsd_out} show qualitative comparisons on randomly selected test samples from each dataset.
Compared with image-only and event-guided baselines, \method restores clearer structural boundaries and preserves more fine details in low-light regions.
The results also contain fewer visible noise artifacts and more natural contrast, especially around edges and textured areas.
The enhanced images exhibit brightness and contrast
closer to the reference images, with improved visibility
in dark regions and controlled enhancement in bright regions.
These visual comparisons indicate that our method helps produce cleaner, sharper, and more faithful results. 

\subsection{Ablation Study}

We examine temporal aggregation on SDE-in and evaluate
event-path contributions and temporal granularity on
SDE-in and SDE-out.
For the aggregation study, all variants retain the same
illumination-residual head and restoration network;
only the aggregation strategy is changed.
Order-Free Accumulation collapses the 32-channel event voxel before feature projection and residual prediction.
Bin-Wise Mean independently projects each temporal bin and then applies order-invariant average pooling.
Ordered ConvGRU corresponds to the complete \term and processes the projected bins sequentially.

Table~\ref{tab:component_ablation} shows that bin-wise
projection improves PSNR from 21.99 to 22.27 dB, while
ordered ConvGRU aggregation further increases it to
22.50 dB.
Relative to early accumulation, the full design improves
PSNR/PSNR* by 0.51/0.42 dB and SSIM by 0.0165,
demonstrating the benefit of preserving bin-level features
and modeling their temporal evolution.

\begin{table}[t]
\centering
\scriptsize
\caption{Ablation of temporal aggregation in \term on SDE-in. All variants use the same residual head and restoration network.}
\label{tab:component_ablation}
\begin{tabular}{lccc}
\toprule
Variant & PSNR & PSNR* & SSIM \\
\midrule
Order-Free Accumulation & 21.99 & 24.58 & 0.7534 \\
Bin-Wise Mean & 22.27 & 24.85 & 0.7650 \\
\rowcolor{LightPurple}
Ordered ConvGRU (\term) & \best{22.50} & \best{25.00} & \best{0.7699} \\
\bottomrule
\end{tabular}
\end{table}

To evaluate the independent contribution of \term, we isolate the two event
pathways in \method. Static event only disables the temporal illumination
residual by setting $\alpha=0$, while retaining the complete 32-channel event voxel,
with its original temporal channel positions, as input
to the restoration network.
Here, ``static'' refers to processing the voxel as
a channel stack without the recurrent illumination
correction provided by \term. TERM only removes the static event input from the
restoration network and allows events to affect the output solely through the
bounded illumination residual. Full \method combines both pathways. All other
model components and training settings remain unchanged.

\begin{table}[t]
\centering
\scriptsize
\caption{Ablation study on the independent contributions of static event
fusion and \term.}
\label{tab:term_isolation_controls}
\resizebox{\columnwidth}{!}{%
\begin{tabular}{l ccc ccc}
\toprule
Variant & \multicolumn{3}{c}{SDE-in}
        & \multicolumn{3}{c}{SDE-out} \\
\cmidrule(lr){2-4}\cmidrule(lr){5-7}
& PSNR & PSNR* & SSIM
& PSNR & PSNR* & SSIM \\
\midrule
TERM only
& 21.03 & 23.14 & 0.6920
& 22.32 & 24.43 & 0.6959 \\

Static event only
& 21.98 & 24.41 & 0.7527
& 21.49 & 25.08 & 0.7159 \\

\rowcolor{LightPurple}
Full \method
& \best{22.50} & \best{25.00} & \best{0.7699}
& \best{22.72} & \best{25.78} & \best{0.7540} \\
\bottomrule
\end{tabular}%
}
\end{table}

As shown in Table~\ref{tab:term_isolation_controls}, neither event pathway
alone matches their combination. Relative to the stronger single-path model
for each metric, full \method improves SDE-in by 0.52 dB in PSNR, 0.59 dB in
PSNR*, and 0.0172 in SSIM. On SDE-out, it improves PSNR by 0.40 dB over TERM
only, and improves PSNR* and SSIM by 0.70 dB and 0.0381 over static event only,
respectively. On SDE-out, TERM only achieves higher PSNR, while
static event fusion yields higher PSNR* and SSIM.
Their combination achieves the best results across
all three metrics, demonstrating the complementary
roles of temporal illumination correction and static
event fusion. These results support the
complementarity of the two pathways: static event fusion provides structural
event evidence directly to the restoration network, whereas \term exploits
ordered photometric dynamics to refine the image-derived illumination
estimate.

We also study the number of temporal bins $K$ in \term.
Table~\ref{tab:k_bins_ablation} compares $K=4$, $K=8$, and $K=16$ on SDE-in and SDE-out under the same model configuration.
A smaller $K$ provides denser events per bin but coarser temporal resolution, while a larger $K$ improves temporal granularity but makes each recurrent input sparser.
The best results are consistently obtained with $K=8$ on all six dataset-metric pairs.
Compared with $K=4$ and $K=16$, $K=8$ improves PSNR by 0.09 dB and 0.35 dB on SDE-in, and by 0.78 dB and 0.39 dB on SDE-out, respectively.
It indicates that $K=8$ provides a better balance between temporal resolution and event density.

\begin{table}[t]
\centering
\scriptsize
\caption{Parameter ablation of the number of temporal bins $K$ in \term on SDE-in and SDE-out.}
\label{tab:k_bins_ablation}
\resizebox{\columnwidth}{!}{%
\begin{tabular}{c ccc ccc}
\toprule
$K$ & \multicolumn{3}{c}{SDE-in} & \multicolumn{3}{c}{SDE-out} \\
\cmidrule(lr){2-4}\cmidrule(lr){5-7}
& PSNR & PSNR* & SSIM & PSNR & PSNR* & SSIM \\
\midrule
4  & 22.41 & 24.58 & 0.7639 & 21.94 & 25.59 & 0.7335 \\
\rowcolor{LightPurple}
8  & \best{22.50} & \best{25.00} & \best{0.7699}
   & \best{22.72} & \best{25.78} & \best{0.7540} \\
16 & 22.15 & 24.79 & 0.7625 & 22.33 & 25.35 & 0.7331 \\
\bottomrule
\end{tabular}%
}
\end{table}

\subsection{Analysis of Temporal Feature Placement}
We further investigate where the temporal feature should be introduced in
the enhancement pipeline. Direct temporal fusion maps the final ConvGRU state
to the event representation of the restoration network through a
zero-initialized $1\times1$ projection, without applying the temporal
illumination residual. Direct + illumination injects the temporal feature
into both the restoration network and the illumination branch. In contrast,
the illumination-branch design uses the temporal state only to predict the
bounded illumination residual, while retaining the original static event
path in the restoration network. All three variants use the same temporal
encoder, restoration backbone, and training settings.

\begin{table}[t]
\centering
\scriptsize
\caption{Ablation study on temporal-feature placement on SDE-in and SDE-out.}
\label{tab:term_placement_controls}
\resizebox{\columnwidth}{!}{%
\begin{tabular}{l ccc ccc}
\toprule
Placement & \multicolumn{3}{c}{SDE-in}
          & \multicolumn{3}{c}{SDE-out} \\
\cmidrule(lr){2-4}\cmidrule(lr){5-7}
& PSNR & PSNR* & SSIM
& PSNR & PSNR* & SSIM \\
\midrule
Direct temporal fusion
& 21.73 & 24.09 & 0.7402
& 22.14 & 25.46 & 0.7431 \\

Direct + illumination
& 22.04 & 24.48 & 0.7553
& 22.49 & 25.48 & 0.7350 \\

\rowcolor{LightPurple}
Illumination branch
& \best{22.50} & \best{25.00} & \best{0.7699}
& \best{22.72} & \best{25.78} & \best{0.7540} \\
\bottomrule
\end{tabular}%
}
\end{table}

As shown in Table~\ref{tab:term_placement_controls},
illumination-branch injection achieves the best results
across all three metrics on both datasets.
Compared with direct temporal fusion, it improves PSNR
by 0.77 dB on SDE-in and 0.58 dB on SDE-out, with
corresponding gains in PSNR* and SSIM.
Adding direct fusion to illumination correction does not
provide further improvements.
These results demonstrate the importance of temporal-feature
placement and support bounded illumination correction
as an effective way to incorporate ordered event dynamics
into low-light restoration.

\subsection{Effect of chronological event order}
To isolate the effect of temporal order, all variants use the same data
splits, preprocessing, restoration backbone, loss functions, and training
schedule, and differ only in the temporal aggregation strategy. Each variant is trained independently from scratch and
evaluated with the same ordering rule used during training.
For random-fixed order, one permutation is shared across
all samples throughout training and testing.
All variants use seed 2333, and their checkpoints are
selected by the same minimum-validation-loss criterion. In the
correct-order setting, the event voxel is divided into $K=8$ bins according
to ascending timestamps, and the same ConvGRU processes them sequentially
from the earliest bin to the latest. The reverse-order variant uses exactly
the same ConvGRU but processes the bins from latest to earliest. The parameter-matched invariant
variant applies a shared projection to each bin, averages the projected
features, and uses an additional convolutional capacity matcher before
predicting the residual. This invariant variant differs from the recurrent
\term by only 48 parameters.

As shown in Table~\ref{tab:order_controls}, the correct chronological order
achieves the best result for every metric on both SDE-in and SDE-out.
Compared with the strongest alternative for each metric, it improves SDE-in
by 0.08 dB in PSNR, 0.26 dB in PSNR*, and 0.0086 in SSIM. The corresponding
improvements on SDE-out are 0.06 dB, 0.25 dB, and 0.0074, respectively.
Reversing or randomly permuting the bins causes larger degradation,
particularly on SDE-out, while the parameter-matched invariant model remains
consistently inferior to the correct-order \term.

The reverse and random-fixed controls retain the same
recurrent architecture, while the invariant alternative
closely matches its parameter count.
The consistent advantage of chronological processing
therefore supports the contribution of event order beyond
aggregation capacity.

The controlled studies examine complementary aspects of
the temporal-residual design. The order controls compare
different event sequences within the same ConvGRU
architecture, the pathway controls evaluate the contribution
of temporal illumination correction alongside static event
fusion, and the placement controls compare photometric
correction with direct restoration fusion.
Together, they support chronological event modeling
through the illumination pathway.

\subsection{Residual Visualization}

\begin{figure}[t]
\centering
{
\setlength{\tabcolsep}{0.5pt}
\setlength{\fboxsep}{0pt}
\setlength{\fboxrule}{0.4pt}
\renewcommand{\arraystretch}{0.95}
\scriptsize
\begin{tabular}{@{}ccccc@{}}
Input & Base $L_0$ & Residual $\Delta L$ & Corrected $L$ & Output \\
\fbox{\includegraphics[width=0.185\linewidth]{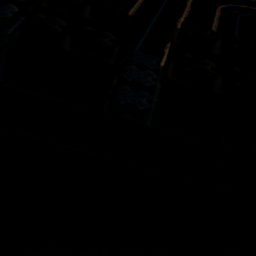}} &
\fbox{\includegraphics[width=0.185\linewidth]{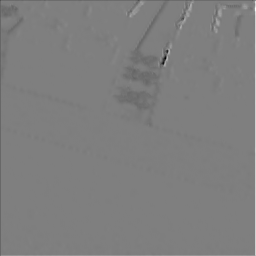}} &
\fbox{\includegraphics[width=0.185\linewidth]{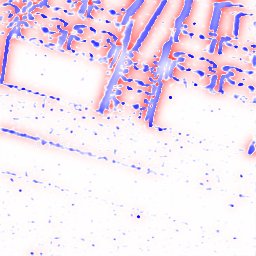}} &
\fbox{\includegraphics[width=0.185\linewidth]{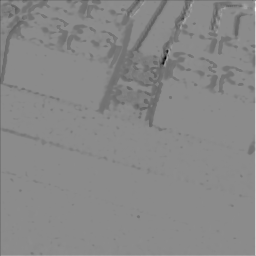}} &
\fbox{\includegraphics[width=0.185\linewidth]{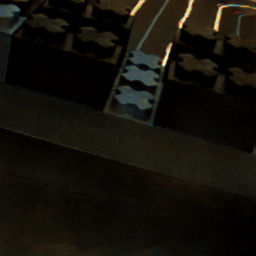}} \\
\fbox{\includegraphics[width=0.185\linewidth]{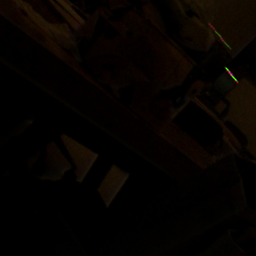}} &
\fbox{\includegraphics[width=0.185\linewidth]{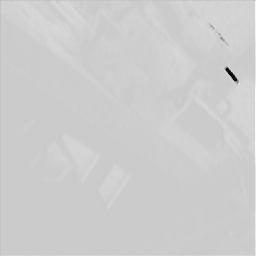}} &
\fbox{\includegraphics[width=0.185\linewidth]{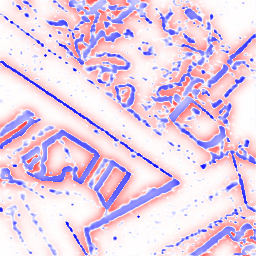}} &
\fbox{\includegraphics[width=0.185\linewidth]{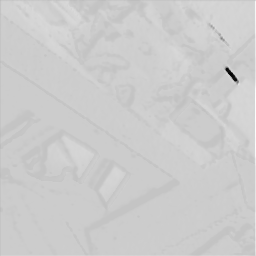}} &
\fbox{\includegraphics[width=0.185\linewidth]{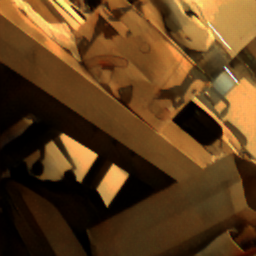}} \\
\fbox{\includegraphics[width=0.185\linewidth]{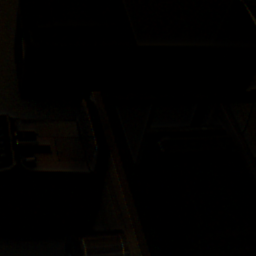}} &
\fbox{\includegraphics[width=0.185\linewidth]{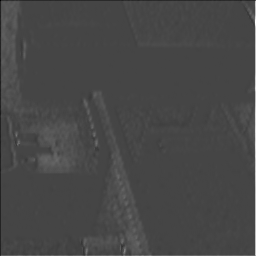}} &
\fbox{\includegraphics[width=0.185\linewidth]{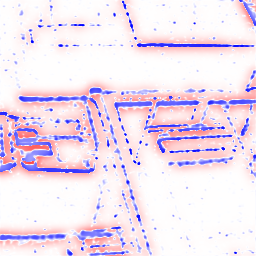}} &
\fbox{\includegraphics[width=0.185\linewidth]{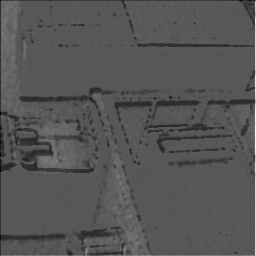}} &
\fbox{\includegraphics[width=0.185\linewidth]{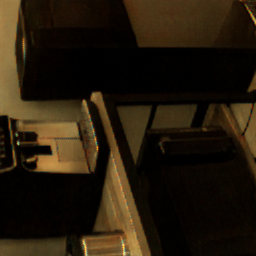}} \\
\end{tabular}
}
\caption{Visualization of learned illumination residuals on three SDE-in samples. From left to right: low-light input, base illumination, residual heatmap predicted from ordered events, corrected illumination, and enhanced output.}
\label{fig:residual_visualization}
\end{figure}

Fig.~\ref{fig:residual_visualization} provides a direct view of how \term modifies the illumination path.
The low-light inputs contain weak contrast and poorly visible structures, while the base illumination maps capture only the coarse exposure distribution.
The learned residuals are spatially non-uniform and concentrate around event-supported edges, texture boundaries, and local brightness transitions.
After adding the residual to the base illumination, the corrected maps preserve the coarse lighting layout while sharpening local photometric details.
The resulting outputs recover clearer scene structures and more balanced illumination.
This behavior is consistent with the ablation results: bin-level features outperform early accumulation, while Ordered ConvGRU further improves restoration through localized temporal illumination correction.

\begin{table}[t]
\centering
\scriptsize
\caption{Ablation study on temporal event order on both datasets.}
\label{tab:order_controls}
\resizebox{\columnwidth}{!}{%
\begin{tabular}{l ccc ccc}
\toprule
Variant & \multicolumn{3}{c}{SDE-in} & \multicolumn{3}{c}{SDE-out} \\
\cmidrule(lr){2-4}\cmidrule(lr){5-7}
& PSNR & PSNR* & SSIM & PSNR & PSNR* & SSIM \\
\midrule
Random-fixed order
& 22.42 & 24.74 & 0.7613
& 22.12 & 25.04 & 0.7203 \\

Reverse order
& 22.02 & 24.39 & 0.7489
& 22.29 & 25.33 & 0.7276 \\

Param.-matched invariant
& 22.13 & 24.40 & 0.7522
& 22.66 & 25.53 & 0.7466 \\

\rowcolor{LightPurple}
Correct order
& \best{22.50} & \best{25.00} & \best{0.7699}
& \best{22.72} & \best{25.78} & \best{0.7540} \\
\bottomrule
\end{tabular}%
}
\end{table}

\subsection{Robustness under Difficult Event Conditions}
\label{subsec:robustness}

We evaluate fixed EvLight and \method checkpoints on all 2,741
SDE-in test samples without retraining, using identical perturbed
event tensors.
Voxel-entry dropout randomly zeros entries, while event noise
adds random signed unit perturbations to selected entries.
Spatial mismatch shifts events along both spatial axes with
zero padding; temporal mismatch circularly shifts the ordered
event groups.
These perturbations test sensitivity to sparsity, noise, and
misalignment without assuming a complete physical sensor-noise model.

\begin{table}[t]
\centering
\scriptsize
\caption{Robustness under representative event perturbations on
all 2,741 SDE-in test samples. Both models receive identical
perturbed events. Spatial shifts apply to both axes; temporal
shifts are measured in event groups.}
\label{tab:robustness_main}
\resizebox{\columnwidth}{!}{%
\begin{tabular}{l ccc ccc}
\toprule
Condition & \multicolumn{3}{c}{EvLight}
          & \multicolumn{3}{c}{\method} \\
\cmidrule(lr){2-4}\cmidrule(lr){5-7}
& PSNR & PSNR* & SSIM
& PSNR & PSNR* & SSIM \\
\midrule
Clean
& 21.98 & 24.41 & 0.7527
& \best{22.50} & \best{25.00} & \best{0.7699} \\

Dropout 30\%
& 22.21 & 24.64 & 0.7586
& \best{22.65} & \best{25.29} & \best{0.7726} \\

Dropout 70\%
& 21.29 & 24.11 & 0.7479
& \best{22.16} & \best{24.82} & \best{0.7628} \\

Noise 5\%
& 20.92 & 22.93 & 0.6454
& \best{21.84} & \best{23.07} & \best{0.6689} \\

Spatial shift 10 px
& 20.88 & 22.60 & 0.6587
& \best{21.09} & \best{22.82} & \best{0.6642} \\

Temporal shift 2
& 21.92 & 24.18 & 0.7379
& \best{22.26} & \best{24.80} & \best{0.7666} \\
\bottomrule
\end{tabular}%
}
\end{table}

Table~\ref{tab:robustness_main} shows that \method outperforms
EvLight across all reported metrics and perturbations,
extending its advantage beyond clean and aligned event inputs.
Spatial mismatch narrows this advantage, highlighting the
importance of accurate RGB--event alignment.

At 30\% dropout, both models slightly improve, potentially
reflecting reduced redundancy or noise as well as input
attenuation from unscaled dropout.
Performance decreases at 70\% dropout, indicating the
importance of retaining informative event evidence.

We further divide the clean test set into event-activity
tertiles based on the sum of absolute voxel responses:
914 sparse-, 914 medium-, and 913 high-activity samples.
The corresponding PSNR gains over EvLight are
0.18, 0.38, and 0.99 dB, showing that the advantage persists
with sparse observations and increases with event activity.
Event activity, however, is not a direct measure of motion magnitude.

Severe spatial mismatch remains challenging.
A 10-pixel shift reduces the PSNR advantage over EvLight
from 0.52 to 0.21 dB and the SSIM advantage from 0.0172
to 0.0055.
Thus, although \method retains an advantage under the tested
perturbations, accurate RGB--event alignment remains important
for achieving its full enhancement benefit.

\begin{table}[t]
\centering
\scriptsize
\caption{End-to-end computational efficiency at
$256\times256$, batch size one, and FP32 precision.}
\label{tab:term_complexity}
\resizebox{\columnwidth}{!}{%
\begin{tabular}{lcccc}
\toprule
Model & Params (M) & FLOPs (G) & Time (ms) & FPS \\
\midrule
EvLight & 22.729 & 386.917 & 61.493 & 16.26 \\
\rowcolor{LightPurple}
\method & 22.747 & 395.107 & 62.315 & 16.05 \\
\bottomrule
\end{tabular}%
}
\end{table}

\subsection{Computational Efficiency}
\label{sec:efficiency}

As shown in Table~\ref{tab:term_complexity}, \term adds
17,505 parameters, corresponding to only 0.077\% of
the EvLight parameter count.
The additional computation is 8.190 GFLOPs (2.12\%),
while end-to-end latency increases by 0.822 ms (1.34\%).
The corresponding throughput changes from 16.26 to
16.05 FPS, showing that the temporal-residual design
largely preserves the inference speed of the complete
enhancement pipeline.

This efficiency is consistent with the organization of
\method: sequential processing is confined to the
temporal event encoder, whose final state predicts
a single illumination correction before restoration.
The restoration network is evaluated once per image,
without being repeated for individual temporal bins.
The framework therefore incorporates within-window
event dynamics with limited additional parameters
and computation. Together with the enhancement results in
Table~\ref{tab:main_results}, these measurements demonstrate
improved restoration quality with limited additional
parameters, computation, and latency.

\section{Conclusion}
\label{sec:conclusion}

We propose \method, a temporal-residual enhancement framework for event-guided low-light image enhancement.
Instead of treating the event voxel as a static multi-channel map, \method preserves ordered temporal bins and uses the Temporal Event Residual Module (\term) to convert short-term event dynamics into a bounded illumination correction.
This design couples ordered event dynamics with Retinex-style illumination refinement and provides spatially adaptive photometric guidance according to brightness-change trends.
Experiments on SDE and SDSD indoor/outdoor benchmarks show strong overall performance against image-only and image-event methods, with the best result on eleven of twelve dataset-metric pairs.
The ablation results further demonstrate that bin-level event features and correct temporal event order both contribute to the effectiveness of the illumination correction.
Future work will further explore the generality of the proposed framework under more diverse low-light conditions.

\bibliographystyle{elsarticle-num}
\bibliography{refs}

\end{document}